\documentclass[10pt]{article}

\usepackage[dvipsnames,table]{xcolor}
\usepackage{todonotes}
\usepackage{mathtools}
\usepackage{ifsym}
\usepackage[breaklinks]{hyperref}

\usepackage{tikz}
\usetikzlibrary{calc,matrix,positioning,shapes,chains}
\usetikzlibrary{arrows}
\usepackage{amsthm}
\usepackage{colortbl}

\newcommand{\clement}[1]{\textbf{\textcolor{Blue}{{[Clement: #1]}}}}

\newcommand{\stefano}[1]{\textbf{\textcolor{Magenta}{{[Stefano: #1]}}}}
\newcommand{\luc}[1]{\textbf{\textcolor{WildStrawberry}{{[Luc: #1]}}}}

\usepackage{xspace}
\newcommand{\method}{\textsc{VisualSynth}\xspace}
\newcommand{\psyche}{\textsc{PSyChe}\xspace}
\newcommand{\tacle}{\textsc{TaCLe}\xspace}

\usepackage{tabularx} 
\usepackage{adjustbox} 
\usepackage{subcaption}
\usepackage{microtype}

\begin{document}
\title{Human--Machine Collaboration for Democratizing Data Science}
\author{Clément Gautrais$^1$, Yann Dauxais$^1$, Stefano Teso$^1$\\Samuel Kolb$^1$, Gust Verbruggen$^1$, Luc De Raedt$^1$}
\date{%
    $^1$KU Leuven, Department of Computer Science, Leuven, Belgium\\%
    \today
}
\maketitle

\begin{abstract}
   Everybody wants to analyse their data, but only few posses the data science expertise to to this. Motivated by this observation we introduce a novel framework and system \method for human-machine collaboration in data science. 
  It wants to democratize data science by allowing users to interact with standard spreadsheet software in order to perform and automate various data analysis tasks ranging from data wrangling,  data selection, clustering, constraint learning, predictive modeling and auto-completion. \method relies on the user providing colored sketches, i.e.,
  coloring parts of the spreadsheet, to partially specify data science tasks, which are then determined and executed using artificial intelligence techniques. 
\end{abstract}

\section{Introduction}

Data science is a cornerstone of current business practices.  A major obstacle to its adoption is that most data analysis techniques are beyond the reach of typical end-users.  Spreadsheets are a prime example of this phenomenon: despite being central in all sorts of data processing pipelines, the functionality necessary for processing and analyzing spreadsheets is hidden behind the high wall of spreadsheet formulas, which most end-users can neither write nor understand~\cite{chambers2010struggling}.  As a result, spreadsheets are often manipulated and analyzed manually.  This increases the chance of making mistakes and prevents scaling beyond small data sets.

Lowering the barrier to entry for specifying and solving data science tasks would help ameliorating these issues.  Making data science tools more accessible would lower the cost of designing data processing pipelines and taking data-driven decisions.  At the same time, accessible data science tools can prevent non-experts from relying on fragile heuristics and improvised solutions.

The question we ask is then: \emph{is it possible to enable non-technical end-users to specify and solve data science tasks that match their needs}?

We provide an initial positive answer based on two key observations.  First, many key data science tasks can be partially specified using \emph{colored sketches} only.  Roughly speaking, a sketch is a collection of entries, rows, or columns appearing in a spreadsheet that are highlighted using one or more colors.  A sketch determines some or all of the parameters of a data science task.  For instance, while clustering rows, color highlighting can be used to indicate that some rows belong to the same cluster (by highlighting them with the same color) or to different clusters (with different colors).  This information acts as a partial specification of the data science task.  The main feature of sketches is that they require little to no technical knowledge on the user's end, and therefore can be easily designed and manipulated by na\"ive end-users~\cite{sarkar2015interactive}.

Second, the data science task determined by a sketch can be solved using automated data science techniques.  In other words, since the specification may be missing one or more parameters, the spreadsheet application takes care of figuring these out automatically.  The output of this step is a candidate solution, e.g., a possible clustering of the target rows.  The other key feature of sketches is that the result of the data science task can also often be presented through color highlighting.  For instance, row clusters can be captured using colors only.

These two observations enable us to design an interactive framework, \method, in which the machine and the end-user collaborate towards designing and solving a data science task compatible with the user's needs.  \method combines two components: an interaction protocol that allows non-technical people to design partial data science task specifications using colored highlighting, and a smart framework for automatically solving a partially specified data science task based on inductive models.  

In contrast to automation frameworks like AutoML~\cite{thornton2013auto,feurer2015efficient}, \method does not assume that the data science task is fixed and known a priori.\footnote{Indeed, \method supports explorative data science, in which the user is not sure about the task to be performed and tries out different manipulations until it finds one that is interesting or useful.  A proper discussion of explorative data science, however, falls outside the scope of this chapter.}  We do not claim that our human-machine interaction strategy is ideal, but we do claim that it is quite minimal and that despite its simplicity, it suffices to guide the system towards producing useful data science results for many central data science tasks, as shown in the remainder of this chapter.

\method only requires the end-user to check the solution and make sure that it is as expected.  This substantially reduces the expertise required of the user: almost everybody can interact using color highlighting and check whether a solution is compatible with his needs.  The bulk of the complexity -- namely figuring out the bits that are missing from the user' specification -- is handled by the machine itself.  The intent of this setup is to combine the respective strengths of end-users, namely their knowledge of the domain at hand, and computers, namely their ability to quickly carry out enormous amounts of computation.

The remainder of this chapter is structured as follows.  In Section \ref{sec:use_case}, we motivate our approach using a concrete use case. 
Section~\ref{sec:sketches} discusses sketches for several core data science tasks, including data wrangling, prediction, clustering, and auto-completion, and details how the sketches define interaction. Section~\ref{sec:sketches} also describes how tasks partially defined by sketches are solved by the machine. The chapter ends with some concluding remarks.

\section{Motivation}
\subsection{Spreadsheets}

Spreadsheets are used by hundreds of millions of users and are as such one of the most common interfaces that people use to interact with data.  The reason for their popularity is their flexibility:
1) spreadsheets are very heterogeneous and can contain arbitrary types of data, including numerical, categorical and textual values;
2) data can be explicitly organized using tables and operated on using formulas;
3) the ``data generating process'' is almost arbitrary, as spreadsheets can be used for anything from accounting to financial analysis to stock management.
Since our goal is enable as many users as possible to perform data science, a natural choice is to bring data science to spreadsheets.

This is very challenging, for two reasons.  First and foremost, the vast majority of spreadsheet users have little or no knowledge about how to perform data science.  While these na\"ive  users might have heard of data science -- at least to some degree -- they are likely not technically skilled:  most spreadsheet users cannot program even one-line spreadsheet formulas, nor design small data processing pipelines~\cite{chambers2010struggling}.  For instance, in the ice cream sales example, the sale manager has little experience with spreadsheet software and can rely on his domain knowledge.  Our goal is to make his knowledge operational using interactive data science tools and thus to enable her to tackle her forecasting problem more accurately.

In order to cater to this audience, \method relies on a visual, concrete and interactive protocol in which the user and the machine collaborate to explore the data and design a data processing pipeline.  The protocol leverages simple and intuitive forms of interaction that require no or little supervision and almost zero technical knowledge.  This is achieved through a combination of interaction and automation.

\subsection{A Motivating Example: Ice Cream Sales}
\label{sec:use_case}






Let us now illustrate interactive data science and \method with a classic use case of naive spreadsheet end-users: auto-completion.  Tackling this use case requires collaboration between the user and the machine to convey the intentions and the knowledge of the user, as shown below.

Imagine that you are a sales manager at an ice cream factory. You have data about past sales and some information about your shops, as shown in Tables~\ref{tab:ice_cream_sales_raw} and \ref{tab:city_properties}, respectively.

A first difficulty is that Table~\ref{tab:ice_cream_sales_raw} is not nicely formatted. A first task is  therefore to wrangle Table~\ref{tab:ice_cream_sales_raw} into a format such as that listed in Table~\ref{tab:ice_cream_sales} that is more amenable to data analysis. 
Through interaction, the \textbf{data wrangling component} can produce the table presented in Table~\ref{tab:ice_cream_sales}.

However, some past sales data are missing. To determine which shops made a profit you need to first obtain an estimate of the missing values.
To produce such  estimates,
you can interact with our system in different ways. First, as the sales manager you know that the profit of a shop depends on the type of ice cream and the characteristics of the city. More precisely, you know that some cities have similar profitability profiles. To convey this knowledge, you can use a coloring scheme to indicate that certain cities belong to the same cluster. This will in turn trigger an \textbf{interactive clustering} process,
which not only allows you to state must-link and cannot-link constraints using colorings but also to correct mistakes that our system might make during the clustering process. 
Once the clustering is deemed correct, the machine stores this information and displays it as a new column in the spreadsheet.  From this enriched data, you can then can ask the machine to provide a first \textbf{estimate of the missing values}. This can be achieved in different ways.


First, as a sales manager you could start filling the missing values yourself. After one or two missing values are filled, the machine can infer that the remaining missing values should also be filled. The machine will thus start suggesting values, which you can then either accept them as is or correct them. Corrections will trigger a new auto-completion loop, with additional constraints expressing that the user corrected some values in the previous iteration. 

Second, you could trigger the auto-completion by indicating   that the machine should fill the missing values. For this, you can use colors to indicate which values should be predicted. Then, human-machine interaction proceeds as described above. Additionally, the machine could provide information about some of the underlying model assumptions. For example, the machine can indicate which columns are used for prediction and you could indicate whether these columns are relevant for predicting profit.

The remainder of this paper introduces some principles for human-machine collaboration in the context of auto-completion and automated data science.  In particular, we identify different levels of interaction,
discuss
how the machine adds user-knowledge in its learning mechanisms, and
elucidates
how different data science tasks fit in our framework.

\begin{table}[hbt]
    \centering
    \begin{minipage}{.495\textwidth}
    \begin{tiny}
    \centering
    \begin{tabular}{|l|l|c|} 
        \hline
        Florence &        &     \\ \hline
        Vanilla  & June   & 610 \\ \hline
                 & July   & 190 \\ \hline
                 & Aug    & 670 \\ \hline
                 & Total  & 1470 \\ \hline
                 & Profit & YES \\ \hline
        Stracciatella & June & 300 \\ \hline
                      & July & 250 \\ \hline
                      & Aug  & 290 \\ \hline
                      & Total & 860 \\ \hline
                      & Profit & NO \\ \hline
        $\ldots$      & $\ldots$ & $\ldots$ \\ \hline
        Milan         &          & \\ \hline
        Chocolate & June     & 430 \\ \hline
                  & July     & 350 \\ \hline
                  & Aug      & ? \\ \hline
                  & Total    & ? \\ \hline
                  & Profit   & ? \\ \hline
    \end{tabular}
    \end{tiny}
    \end{minipage}
    \begin{minipage}{.495\textwidth}
    \begin{tiny}
    \centering
    \begin{tabular}{|l|l|l|l|}
    \hline
    City  & Touristic & Weather & Country \\
    \hline
    \hline
    Florence & High & Hot & IT \\
    \hline
    Stockholm & High & Cold & SE \\
    \hline
    Copenhagen & High & Cold & DK\\
    \hline
    Berlin     & Very High & Mild & DE \\
    \hline
    Aachen     & Low & Mild & DE \\
    \hline
    Brussels     & Medium & Mild & BE\\
    \hline
    Milan     & Medium & Hot & IT \\
    \hline
    \end{tabular}
    \end{tiny}
    \end{minipage}
\caption{Left: \label{tab:ice_cream_sales_raw} Spreadsheet with ice cream sale numbers.  The ``?'' values are are missing. The ``?'' values are are missing. Right: \label{tab:city_properties} Spreadsheet containing properties of shops}
\end{table}

\begin{table}[hbt]
\begin{tiny}
    \centering
     \begin{tabular}{|l|l|c|c|c|c|c|} 
    \hline
    Type & City & June & July & Aug & Total & Profit\\
    \hline\hline
    Vanilla & Florence & 610 & 190 & 670 & 1470 & YES \\
    \hline
    Banana & Stockholm & 170 & 690 & 520 & 1380 & YES \\
    \hline
    Chocolate & Copenhagen & 560 & 320 & 140 & 1020 & YES \\
    \hline
    Banana & Berlin &  610 & 640 & 320 & 1570 & NO \\
    \hline
    Stracciatella & Florence & 300 & 270 & 290 & 860 & NO \\
    \hline
    Chocolate & Milan & 430 & 350 & ? & ? & ? \\
    \hline
    Banana & Aachen & 250 & 650 & ? & ? & ? \\
    \hline
    Chocolate & Brussels & 210 & 280 & ? & ? & ? \\
    \hline
    \end{tabular}
\caption{\label{tab:ice_cream_sales} Spreadsheet with ice cream sale numbers.}
\end{tiny}
\end{table}

\section{Data Science Sketches}
\label{sec:sketches}

We now introduce  the interaction strategy of \method, our framework for interactive data science.

Given a spreadsheet, a \emph{sketch} is simply a set of colors (aka \emph{coloring}) applied to one or more rows, columns, or cells appearing in the spreadsheet.  The key idea is that, the colors partially define the parameters (e.g., the type, inputs, and outputs) of a data science task.  Hence, taken together, the sketch and the spreadsheet can be mapped onto a very concrete \emph{data science task} (e.g., a clustering task), which can then be solved and and whose results (e.g., a set of clusters) can be filled into or appended to the original spreadsheet, yielding an extended spreadsheet.  This idea is captured in the following schema:
$$\left .\begin{matrix} \mbox{spreadsheet}\\
+\\
\mbox{sketch}\\\end{matrix} \right \} \rightarrow \begin{matrix}\mbox{}\\
\mbox{data science task}\\\mbox{}\end{matrix} \rightarrow \left .\begin{matrix} \mbox{spreadsheet}\\
+\\
\mbox{model}\\\end{matrix} \right \} \rightarrow \begin{matrix}\mbox{}\\
\mbox{new spreadsheet}\\\mbox{}\end{matrix}$$
When explaining the different components of  \method we shall adhere to the above scheme, i.e., our examples and figures will consist of four components: 1) the input sketch and spreadsheet, 2) the data science problem specification, 3) the model, and 4) the resulting spreadsheet.

The above scheme 
 is in line with the closure property of  databases and inductive database \cite{imielinski1996database,de2002perspective}.
 For relational databases, both the inputs and the results of a query are relations, which guarantees
 that the results of one query can be used as the input for the next.
 In a similar vein, in our setting, the inputs as well as the
result of each operation (or data science task) are tables in a spreadsheet.  The closure property  guarantees that 
further analysis is possible after each data science task. 

\method is an example of user-guided interaction that enables the user to convey her intentions by interacting using visual cues.  Indeed, the sketches are supplied by and end-user and are gradually refined in an interactive fashion -- thus adapting the data science task itself -- until the user is satisfied with the result.

Next, we illustrate this interaction protocol using a number of key data science tasks, namely data wrangling, concept learning, prediction, clustering, constraint learning, and auto-completion.

\subsection{Data Wrangling}
\label{subsec:wrangling}


Wrangling is the task of transforming data in the right format for downstream data science tasks. Coloring cells has already been used to help automated wranglers transform data in a format desired by a user \cite{verbruggen2018automatically}. The user has to indicate which cells belong to the same row by coloring them using the same color.
A \emph{wrangling sketch} is therefore a set of colored cells, where each color defines a partial example of the expected wrangling result and imposes a constraint on the output, i.e., that the partial example should be mapped onto a single row into the target spreadsheet.

A commonly used paradigm for data wrangling is programming by example (PBE), in which a language of transformations $\mathcal{L}$ is defined and the wrangler searches for a program $P \in \mathcal{L}$ that maps the input examples to the corresponding outputs. 
In the context of \method, \textbf{given} a wrangling sketch and a spreadsheet, the goal is to \textbf{find} a program that transforms the spreadsheet in such a way that cells with the same color end up in the same row, and no row can contain cells with multiple colors.

An example is shown in Figure~\ref{tbl:wrangling_input}. The data is clearly not in a suitable format for analysis and a novice user might not be able to efficiently transform it. From a small number of colored cells---the wrangling sketch---the synthesizer is able to learn the program described in Figure~\ref{tbl:wrangling_program}. This program yields the desired table from Figure~\ref{tbl:wrangling_output} when applied on the input table.

Finding such a program is a form of \emph{predictive program synthesis}. The desired solution is not known in explicit form,   but the wrangling sketch imposes a constraint that the solution should at least satisfy. Additionally, syntactic and semantic properties of the elements in rows and columns are used for heuristically determining the quality of candidate solutions.

In addition to defining constraints on the output, the wrangling sketch can be used to define heuristics for improving the search for a correct program. The relative positions of cells in the same or different colors allow one to impose a strong syntactic bias on the program synthesizer. For example, two consecutive columns with the same number of vertically adjacent cells of the same color are very good candidates for a pivot transformation, as in Table~\ref{tab:ice_cream_wrangling}. A greedy beam search that interleaves heuristically selecting transformations and evaluating the results of these transformations was used in \cite{verbruggen2018automatically} to quickly find spreadsheet transformation programs.

\begin{table}[hbt]
     \centering
     \begin{subtable}[t]{.45\textwidth}
     \centering
     \resizebox{!}{2.5cm}{
     \begin{tabular}{|l|l|c|} 
         \hline
         \cellcolor{blue!25}Florence &        &     \\ \hline
         \cellcolor{blue!25}Vanilla  & June   & \cellcolor{blue!25}610 \\ \hline
                  & July   & \cellcolor{blue!25}190 \\ \hline
                  & Aug    & \cellcolor{blue!25}670 \\ \hline
                  & Total  & \cellcolor{blue!25}1470 \\ \hline
                  & Profit & \cellcolor{blue!25}YES \\ \hline
         Stracciatella & June & 300 \\ \hline
                       & July & 250 \\ \hline
                       & Aug  & 290 \\ \hline
                       & Total & 860 \\ \hline
                       & Profit & NO \\ \hline
         $\ldots$      & $\ldots$ & $\ldots$ \\ \hline
         Milan         &          & \\ \hline
         Chocolate & \cellcolor{red!25}June     & 430 \\ \hline
                   & \cellcolor{red!25}July     & 350 \\ \hline
                   & \cellcolor{red!25}Aug      & ? \\ \hline
                   & \cellcolor{red!25}Total    & ? \\ \hline
                   & \cellcolor{red!25}Profit   & ? \\ \hline
     \end{tabular}}
     \caption{Input data and wrangling sketch where each color indicates cells that should end up in the same row.}
     \label{tbl:wrangling_input}
     \end{subtable}
     \begin{subtable}[t]{.53\textwidth}
     \centering
         \begin{tabularx}{\textwidth}{X}
         	\textbf{Given} the blue and red colorings,\\ 
         	a spreadsheet, \\
         	 a language in ${\cal L}$ in which to express programs,\\
         	\textbf{Find} a wrangling program $P \in \mathcal{L}$\\
         	such that blue cells end up in a single row,\\
         	and red cells in another single row.
         \end{tabularx}
          \caption{Wrangling problem statement.}
          \label{tbl:wrangling_problem_setting}
     \end{subtable}
     
     \begin{subtable}[b]{.35\textwidth}
     \centering
         \begin{tabularx}{\textwidth}{X}
         	\textsf{split} column 1 into two columns based on having a value in columns 2 and 3\\
             \textsf{forward fill} column 1 \\
             \textsf{forward fill} column 2 \\
    		 \textsf{pivot} columns 3 and 4 \\
         \end{tabularx}
          \caption{High-level description of the transformation program, the model of the data wrangling task. See Table~\ref{tab:wrangling_transformations} for examples of wrangling functions.}
          \label{tbl:wrangling_program}
     \end{subtable}
     \begin{subtable}[b]{.6\textwidth}
     \centering
     \resizebox{\textwidth}{!}{
     \begin{tabular}{|l|l|c|c|c|c|c|} 
    \hline
     &  & \cellcolor{red!25}June & \cellcolor{red!25}July & \cellcolor{red!25}Aug & \cellcolor{red!25}Total & \cellcolor{red!25}Profit\\
    \hline
    \cellcolor{blue!25}Vanilla & \cellcolor{blue!25}Florence & \cellcolor{blue!25}610 & \cellcolor{blue!25}190 & \cellcolor{blue!25}670 & \cellcolor{blue!25}1470 & \cellcolor{blue!25}YES \\
    \hline
    Banana & Stockholm & 170 & 690 & 520 & 1380 & YES \\
    \hline
    Chocolate & Copenhagen & 560 & 320 & 140 & 1020 & YES \\
    \hline
    Banana & Berlin &  610 & 640 & 320 & 1570 & NO \\
    \hline
    Stracciatella & Florence & 300 & 270 & 290 & 860 & NO \\
    \hline
    Chocolate & Milan & 430 & 350 & ? & ? & ? \\
    \hline
    Banana & Aachen & 250 & 650 & ? & ? & ? \\
    \hline
    Chocolate & Brussels & 210 & 280 & ? & ? & ? \\
    \hline
    \end{tabular}
     }
     \caption{Expected output of the wrangling task.}
     \label{tbl:wrangling_output}
     \end{subtable}
 \caption{Input and expected output of the wrangling task.}
 \label{tab:ice_cream_wrangling} 
 \end{table}
 
 \begin{table}[h]
\centering
\begin{tikzpicture}[
     node distance = 8mm and 30mm,
every matrix/.style = {matrix of nodes,
                      nodes={draw},
                      column sep=-\pgflinewidth,
                      row sep=-\pgflinewidth}
                        ]
\matrix (m1) {
|[draw,fill=green!20]| & |[draw]| & |[draw]| & \\
|[draw,fill=red!20]| & |[draw]| & |[draw]| & \\
|[draw,fill=red!20]| & |[draw]| & |[draw]| & \\
};

\matrix (m2) [right=of m1]{
|[draw,fill=green!20]| & |[draw]| & |[draw]|& |[draw]| & \\
|[draw]| & |[draw,fill=red!20]| & |[draw]|& |[draw]| & \\
|[draw]| & |[draw,fill=red!20]| & |[draw]|& |[draw]| & \\
};

\draw[->] (m1) -- node [above] {\textbf{ Split(1)}} (m2);

\matrix (m3) [below=of m1] {
|[draw,fill=green!20]| & |[draw,fill=red!20]| \\
|[draw,fill=green!20]| & |[draw]| \\
|[draw,fill=green!20]| & |[draw]| \\
|[draw,fill=green!20]| & |[draw,fill=red!60]| \\
|[draw,fill=green!20]| & |[draw]| \\
|[draw,fill=green!20]| & |[draw]| \\
};

\matrix (m4) [right=of m3]{
|[draw,fill=green!20]| & |[draw,fill=red!20]| \\
|[draw,fill=green!20]| & |[draw,fill=red!20]| \\
|[draw,fill=green!20]| & |[draw,fill=red!20]| \\
|[draw,fill=green!20]| & |[draw,fill=red!60]| \\
|[draw,fill=green!20]| & |[draw,fill=red!60]| \\
|[draw,fill=green!20]| & |[draw,fill=red!60]| \\
};

\draw[->] (m3) -- node [above] {\textbf{ Forward Fill(2)}} (m4);

\matrix (m5) [below=of m3] {
|[draw,fill=green!20]| & |[draw,fill=red!20]| \\
|[draw,fill=green!20]| & |[draw,fill=red!40]| \\
|[draw,fill=green!60]| & |[draw,fill=red!60]| \\
|[draw,fill=green!60]| & |[draw,fill=red!80]| \\
};

\matrix (m6) [right=of m5]{
|[draw,fill=green!20]| & |[draw,fill=green!60]|\\
 |[draw,fill=red!20]| & |[draw,fill=red!60]|\\
|[draw,fill=red!40]| & |[draw,fill=red!80]|\\
};

\draw[->] (m5) -- node [above] {\textbf{ Pivot(1,2)}} (m6);
\end{tikzpicture}
\caption{\label{tab:wrangling_transformations}Examples of wrangling functions. \textbf{Split} creates a new column for each value of a given column. \textbf{Forward fill} fills missing values in a column with the value directly above it. \textbf{Pivot} uses unique values of a column as a new set of columns.}
\end{table}

\subsection{Data Selection}

\begin{table}[h!]
\tiny
\centering
\begin{subtable}{\textwidth}
\centering
    \begin{tabular}{|l|l|c|c|c|c|c|} 
    \hline
    Type & City & June & July & Aug & Total & Profit\\
    \hline\hline
    \cellcolor{blue!25}Vanilla & \cellcolor{blue!25} Florence & \cellcolor{blue!25} 610 & \cellcolor{blue!25} 190 & \cellcolor{blue!25} 670 & 1470 & \cellcolor{blue!25} YES \\
    \hline
    Banana & Stockholm & 170 & 690 & 520 & 1380 & YES \\
    \hline
    \cellcolor{blue!25} Chocolate & \cellcolor{blue!10} Copenhagen & \cellcolor{blue!10} 560 & \cellcolor{blue!10} 320 & \cellcolor{blue!10} 140 & 1020 & \cellcolor{blue!10} YES \\
    \hline
    Banana & Berlin &  610 & 640 & 320 & 1570 & NO \\
    \hline
    Stracciatella & Florence & 300 & 270 & 290 & 860 & NO \\
    \hline
    \cellcolor{blue!25} Chocolate & \cellcolor{blue!25} Milan & \cellcolor{blue!10} 430 & \cellcolor{blue!10} 350 & \cellcolor{blue!10} ? & ? & \cellcolor{blue!10} ? \\
    \hline
    Banana & Aachen & 250 & 650 & ? & ? & ? \\
    \hline
    \cellcolor{blue!25} Chocolate & \cellcolor{blue!25} Brussels & \cellcolor{blue!10} 210 & \cellcolor{blue!10} 280 & \cellcolor{blue!10} ? & ? & \cellcolor{blue!10} ? \\
    \hline
    \end{tabular}

\vspace*{0.2cm}

    \begin{tabular}{|l|l|c|c|c|} 
    \hline
    Type & City & ProviderID & Price & Quality\\
    \hline\hline
    \cellcolor{blue!25} Vanilla & \cellcolor{blue!25} Florence & 1 & \cellcolor{blue!25} Cheap & \cellcolor{blue!25} Bad \\
    \hline
    \cellcolor{blue!25} Vanilla & \cellcolor{blue!10} Florence & 2 & \cellcolor{blue!25} Regular & \cellcolor{blue!10} Good \\
    \hline
    Stracciatella & Florence & 1 & Regular & Great \\
    \hline
    \cellcolor{blue!25} Chocolate & \cellcolor{blue!10} Copenhagen & 3 & \cellcolor{blue!10} Cheap & \cellcolor{blue!10} Good \\
    \hline
    Chocolate & Milan & 4 & Regular & Good \\
    \hline
    \cellcolor{red!10} Chocolate & \cellcolor{red!10} Milan & 5 & \cellcolor{red!25} Expensive & \cellcolor{red!10} Great \\
    \hline
    Chocolate & Brussels & 6 & Regular & Good \\
    \hline
    \cellcolor{red!10} Chocolate & \cellcolor{red!10} Brussels & 6 & \cellcolor{red!25} Expensive & \cellcolor{red!10} Good \\
    \hline
    \end{tabular}
\caption{\label{tab:data_selection_input} Input tables describing ice cream sales and providers and containing colored examples. 
    The cells colored in blue and red are the relevant and irrelevant examples, respectively. 
    The cells colored in lighter gradient are the extension of the partial examples. 
}
\end{subtable}

\vspace*{0.2cm}
\begin{subtable}{0.3\textwidth}
\small
 \textbf{Given} positive (blue)  and negative (pink) tuples in a spreadsheet,\\
 the schema of the tables,\\
\textbf{Find} one or more queries that together cover all positives and none of the negatives. 
     \caption{\label{tab:data_selection_given} Problem statement}
\end{subtable}
\hfill
\begin{subtable}{0.69\textwidth}
\small
     \begin{verbatim}
   ?- sales(I0, Type, City, June, July, Aug, 'YES'),
    provider(I1, Type, City, 'Cheap', Quality). 
    
   ?- sales(I0, Type, City, June, July, Aug, Profit),
    provider(I1, Type, City, 'Regular', 'Good').
    \end{verbatim}
    \caption{\label{tab:data_selection_model} Queries describing which rows are positive.}
\end{subtable}

\begin{subtable}{\textwidth}
\centering
    \begin{tabular}{|l|l|c|c|c|c|c|} 
    \hline
    Type & City & June & July & Aug & Total & Profit\\
    \hline\hline
    \cellcolor{blue!25} Vanilla & \cellcolor{blue!25} Florence & \cellcolor{blue!25} 610 & \cellcolor{blue!25} 190 & \cellcolor{blue!25} 670 & 1470 & \cellcolor{blue!25} YES \\
    \hline
    Banana & Stockholm & 170 & 690 & 520 & 1380 & YES \\
    \hline
    \cellcolor{blue!25} Chocolate & \cellcolor{blue!25} Copenhagen & \cellcolor{blue!25} 560 & \cellcolor{blue!25} 320 & \cellcolor{blue!25} 140 & 1020 & \cellcolor{blue!25} YES \\
    \hline
    Banana & Berlin &  610 & 640 & 320 & 1570 & NO \\
    \hline
    Stracciatella & Florence & 300 & 270 & 290 & 860 & NO \\
    \hline
    \cellcolor{blue!25} Chocolate & \cellcolor{blue!25} Milan & \cellcolor{blue!25} 430 & \cellcolor{blue!25} 350 & \cellcolor{blue!25} ? & ? & \cellcolor{blue!25} ? \\
    \hline
    Banana & Aachen & 250 & 650 & ? & ? & ? \\
    \hline
    \cellcolor{blue!25} Chocolate & \cellcolor{blue!25} Brussels & \cellcolor{blue!25} 210 & \cellcolor{blue!25} 280 & \cellcolor{blue!25} ? & ? & \cellcolor{blue!25} ? \\
    \hline
    \end{tabular}
    
    \vspace*{0.2cm}
    
        \begin{tabular}{|l|l|c|c|c|} 
    \hline
    Type & City & ProviderID & Price & Quality\\
    \hline\hline
    \cellcolor{blue!25} Vanilla & \cellcolor{blue!25} Florence & 1 & \cellcolor{blue!25} Cheap & \cellcolor{blue!25} Bad \\
    \hline
    \cellcolor{blue!25} Vanilla & \cellcolor{blue!25} Florence & 2 & \cellcolor{blue!25} Regular & \cellcolor{blue!25} Good \\
    \hline
    Stracciatella & Florence & 1 & Regular & Great \\
    \hline
    \cellcolor{blue!25} Chocolate & \cellcolor{blue!25} Copenhagen & 3 & \cellcolor{blue!25} Cheap & \cellcolor{blue!25} Good \\
    \hline
    \cellcolor{blue!25} Chocolate & \cellcolor{blue!25} Milan & 4 & \cellcolor{blue!25} Regular & \cellcolor{blue!25} Good \\
    \hline
    Chocolate & Milan & 5 & \cellcolor{red!25} Expensive & Great \\
    \hline
    \cellcolor{blue!25} Chocolate & \cellcolor{blue!25} Brussels & 6 & \cellcolor{blue!25} Regular & \cellcolor{blue!25} Good \\
    \hline
    Chocolate & Brussels & 6 & \cellcolor{red!25} Expensive & Good \\
    \hline
    \end{tabular}
    \caption{\label{tab:data_selection_output} Output tables corresponding to the input tables in which the relevant colors are included.}
\end{subtable}
    \caption{\label{fig:data_selection} Input, model and output of the data selection ice cream factory example. 
    The input and output are sets of colored cells from a set of tables and the output is a set of rules representing the set of colored cells to be output.}
\end{table}

Selecting the right data to analyse is one of the essential steps in  data science processes \cite{fayyad1996kdd}. Within \method we view this as the task to extract a subset of subtables from the original spreadsheet. 
This is often a necessary step  before  the machine learning methods (proposed in the following sections) can be applied.

Consider Table \ref{fig:data_selection} as a running example. 
This table can be decomposed as 1) the dataset given as input \ref{tab:data_selection_input}, 2) the problem statement \ref{tab:data_selection_given}, 3) an example of model \ref{tab:data_selection_model} used to represent the selection and 3) the dataset returned as output \ref{tab:data_selection_output}.   
The dataset is represented by two spreadsheet tables. 
The \textit{sales} table gathers the log of each ice cream benefits in each city and the \textit{provider} table gathers the information about the ice cream providers in each city with a discrete evaluation of the price and quality of their products. 

As an example for data selection, if the user wants to predict the missing values for the Chocolate flavour, she could want to predict these using only the known values for Chocolate and Vanilla without considering Banana and Stracciatella based on her knowledge of the ice cream market. 
However, it would be hard for a non expert spreadsheet user to perform the selection by hand. Therefore, the set of rows to be used could
be induced from a set of examples using a sketch. 
In a data selection sketch, the user can indicate desirable examples 
by coloring them in blue, and unwanted or irrelevant ones by coloring them in another color  (say pink). 
The goal of data selection is then to learn which part of the spreadsheet to retain. 
   The model that is learned will consist of queries that, when performed
on the spreadsheet, returns the desired selection of the data. 


As illustrated on both tables with the columns \textit{Total} and \textit{ProviderID}, if a column or a table does not contain any colored cell, this column or table will not appear in the final selection.  This is an intuitive way to represent the projection operator from relational algebra. It ensures that the user can specify partial examples, that is, examples that do not extend over all colored columns or tables.
These partial examples are then automatically extended over the remaining columns as to consider the full rows in the relevant tables.
An example of such a coloring extension is illustrated on the input tables with lighter blue and red for positive and negative examples, respectively. 

The data selection sketch can, thus, be decomposed in two steps. 
First, the coloring of the partial examples is extended to the obtain complete examples. Each example corresponds to a set of rows (or tuples) that can belong to multiple tables. 
Second, the  examples are generalized into queries that should capture the concept underlying the data selection process. Thus, the data selection task can  be formalized as an inductive logic programming or logical and relational learning  problem \cite{de2008logical,muggleton1994inductive} such that:
{\bf Given} 1) a set of tables in a spreadsheet, 2) a set of partial examples in two colors (representing positive and negative examples), 3) the schema of the tables in the spreadsheet, {\bf Find}
one more relational queries whose answers cover all positive examples, and none of the negatives tuples. 
The resulting queries are then run on the tables in the spreadsheet, and all rows that satisfy the query are colored positively.

It will be assumed that we possess some information about the underlying relational schema, in particular, the foreign key relations need to be known. These can be induced by learning systems such as Tacle \cite{kolb2017learning}, which is explained in more detail below.

The use of colors to induce queries was  already considered in a  database setting \cite{bonifati2016learning}. 
However, the focus was on learning the definition of a single relation, not on performing data selection across multiple tables as we do. 
Furthermore, partial examples, which provide the user with extra flexibility, was not  considered. 
\subsubsection{Processing the data} 
The first step is to extend the input coloring
 of Table \ref{tab:data_selection_input} 
 into a set of examples. 
  This process starts from the template and uses  the foreign key relations to indicate the joins. For our running example, the template query is:
 	\begin{verbatim}
?-sales(I0, Type, City, June, July, Aug, Profit),	
	provider(I1, Type, City, Price, Quality).
\end{verbatim}

To select the examples, we start by 
detecting which rows contain at least one color,
and we expand these into the sets of facts we note $Sales^+$ and $Provider^+$, and two other sets matching the irrelevant rows that we note  $Sales^-$ and $Provider^-$, respectively. 
Furthermore, we omit the columns that do not contain any color, as they are deemed irrelevant.

The next step is then to construct the positive examples by taking every ground atom  from 
one of the positive sets $Sales^+$ and $Provider^+$ and unifying it with the corresponding atom for the same predicate in the template. The set of all answers to the query constitutes an example. For instance, the first tuple in the  $Sales^+$ table (having $Type=Vannila$ and $City=Florence$) would yield
the example consisting of that tuple and the 
first two tuples of the $Provider$ table. 
The negative tables are not expanded, they are only used to prune candidate generalizations.

\subsubsection{Relational rule learning}
With this setup, we can now define the inductive logic programming problem \cite{de2008logical}. \textbf{Given} a set of positive examples (where each example is a set of facts), a set of negative examples (the tuples in the negative set),
and the relational structure of the spreadsheet, \textbf{Find} a set of queries that cover all the positive examples  and none of the negative tuples.  
Such queries can in principle be induced using standard relational learners such as 
 GOLEM \cite{muggleton1990efficient} and FOIL \cite{quinlan1990learning}. 

What is used in \method is a simplified GOLEM; \method uses Plotkin's least general generalization (lgg) operator \cite{de2008logical,plotkin1970note} 
together with GOLEM's search strategy.  The lgg operator takes two examples and produces a generalized set of facts that can serve as the query. More specifically, consider the  example related to $Type=Vannila$ and $City=Florence$
and that related to $Type=Chocolate$ and $City=Copenhagen$.
The resulting $lgg$ would be
\begin{verbatim}
?- sales(I0, Type, City, June, July, Aug, "YES"), 
provider(I1, Type, City, "Cheap", Quality),
provider(I2, Type, City, Price, "Good").
\end{verbatim}
The strategy followed by GOLEM  that we adopt here is to  sample positive examples, compute their lgg, verify that the lgg does not cover negative tuples, and if so replace the positive examples (and other positives that are subsumed) by the lgg. This process is continued, until further generalizations yields queries that cover negative tuples and are too general. b 
Applying this strategy to our example yields the two queries shown in Table \ref{tab:data_selection_model}.c.
Evaluating these queries on the original tables results in  Table \ref{tab:data_selection_model}.d.

Finally, the result of these rules which represent the rows to color can be easily matched with the initial template that represents the columns to color and thus, output the result set of colored cells. 

\subsection{Clustering}

Clustering is the task of grouping data in different coherent clusters and is a building block of typical data processing pipelines~\cite{xu2005survey}.  In our use case, we use clustering not only as a way to learn clusters in the data, but also as a way to generate new features. Through clustering, a user can express some of her knowledge explicitly and this knowledge can then be used for future data science steps, such as predicting a missing value.

Since clustering is ill-defined, recent developments in this area enable the machine to interactively elicit knowledge from the end-user so as to guide the clustering towards the user's needs, cf.~\cite{van2018cobras}.  In the simplest case, the machine iteratively presents pairs of (appropriately chosen) examples to the user and asks whether they belong to the same cluster or not. The user's feedback is then translated into pairwise constraints, namely \emph{must-link} and \emph{cannot-link} constraints, which are then used to bias the clustering process according to the elicited knowledge~\cite{wagstaff2001constrained,van2017cobra}.

Building on top of such techniques, colored sketches can be used to implement the interaction: the user colors (a few) objects belonging to the same cluster using the same color. Hence, items highlighted with the same color belong to the same cluster. The sketch therefore consists of a set of such colorings, each identifying examples from a given cluster.  An example sketch is given in Table~\ref{tab:city_properties_sketch}. In this example, the user colored a few rows to indicate that
the city shops in Milan and Florence (both colored in green) should belong to the same cluster, while Berlin and Seville belong to a different cluster (colored in blue). The extra empty column at the end of each table contains the resulting clustering. Although incomplete, this information  often suffices to guide the clustering algorithm towards a clustering compatible with the user's requirements~\cite{van2017cobra}.

\subsubsection{Problem setting}
In section~\ref{subsec:wrangling}, we presented how data wrangling can map an example to a single row of a table. Hence, we consider that an example in the clustering is a table row.
From this observation and the sketch described in the previous paragraph, we get the problem setting for clustering:
\textbf{Given} a set of set of colored rows and a set of uncolored rows \textbf{find} a cluster assignment for all rows such that rows in the same colored set belong to the same cluster and no rows in different sets belong to the same cluster,  or equivalently: \textbf{find} a cluster assignment for all rows such that rows in the same colored set belong to the same cluster and the number of clusters is equal to the number of colors.

\subsubsection{Finding a cluster assignment}

Current techniques to solve the above problem statement typically start from a partial cluster assignment where all examples in the same color set are in the same cluster.
This can be achieved by using clustering algorithms using \textit{must-link} and \textit{cannot-link} constraints~\cite{wagstaff2001constrained,basu2004active,van2017cobra}.
\textit{Must-link} constraints are enforced between examples of the same color set, while \textit{cannot-link} constraints are enforced between examples from different color sets.
Then, non-colored examples have to be assigned according to a learned distance metric~\cite{xing2003distance}, or generalizations of existing (partial) clusters.


The resulting cluster assignment is mapped back into a set of colored rows, as depicted in Table \ref{tab:city_properties_sketch_cluster1}.
The user can then modify the resulting cluster assignment by adding new colors or by putting existing color on previously colorless rows.
Iterative refinements of the cluster assignments and of the sketch are then performed, as the user is unlikely to be able to fix all parameters of the clustering task through a single interaction.

\begin{table}[hbt]
\centering
\begin{subtable}[t]{.6\textwidth}

\begin{tiny}
\centering
    \begin{tabular}{|l||l|l|l|l|l|}
    \hline
        &City  & Touristic & Weather & Nat &\\
       \hline
       \hline
    \rowcolor{LimeGreen}
       1&Florence & High & Hot & IT  &\\
       \hline
       \rowcolor{Lavender}
       2&Stockholm & High & Cold & SE & \\
       \hline
       3&Copenhagen & High & Cold & DK &\\
       \hline
       \rowcolor{SkyBlue}
    4&Berlin     & Very High & Mild & DE & \\
    \hline
    5&Aachen     & Low & Mild & DE & \\
    \hline
    \rowcolor{Lavender}
    6&Brussels     & Medium & Mild & BE & \\
    \hline
     \rowcolor{LimeGreen}
    7&Milan     & Medium & Hot & IT & \\
    \hline
    8&Munich & Medium & Mild & DE & \\
       \hline
      9& Paris & Very High & Mild & FR & \\
       \hline
       10&Turin & High & Hot & IT & \\
       \hline
       \rowcolor{SkyBlue}
       11&Seville & High & Hot & ES & \\
       \hline
       12&Valencia & High & Hot & ES & \\
       \hline
    \end{tabular}
    \end{tiny}
    \caption{\label{tab:city_properties_sketch}Sketch for a clustering task. Rows of the same color belong to the same cluster}
    \end{subtable}
    \begin{subtable}[t]{.3\textwidth}
    \begin{scriptsize}
    \begin{tabularx}{\textwidth}{X}
    \textbf{Given} the green, pink and blue examples and the constraints in Table~\ref{tab:clustering_constraints},\\
    \textbf{find} a cluster assignment that satisfies the constraints.
         \end{tabularx}
         \end{scriptsize}
         \caption{Problem setting of the clustering task}
         \end{subtable}
    
    \begin{subtable}[b]{.40\textwidth}
    \begin{scriptsize}
    \begin{tabularx}{\textwidth}{X}
    mustlink(1, 7), mustlink(2, 6), \\
    mustlink(4, 11), cannotlink(1, 2),\\
    cannotlink(1, 6),cannotlink(1, 4),\\
    cannotlink(1, 11),cannotlink(7, 2),\\
    cannotlink(7, 6),cannotlink(7, 4),\\
    cannotlink(7, 11),cannotlink(2, 4),\\
    cannotlink(2, 11),cannotlink(6, 4),\\
    cannotlink(6, 11)\\
         \end{tabularx}
         \end{scriptsize}
         \caption{\label{tab:clustering_constraints}Constraints passed to the clustering algorithm. Arguments are row number, starting from 1.}
         \end{subtable}
    \begin{subtable}[b]{.59\textwidth}
    \begin{tiny}
    \centering
    \begin{tabular}{|l||l|l|l|l|l|}
    \hline
       &City  & Touristic & Weather & Nat & Cluster\\
       \hline
       \hline
    \rowcolor{LimeGreen}
       1&Florence & High & Hot & IT & \\
       \hline
       \rowcolor{Lavender}
       2&Stockholm & High & Cold & SE & \\
       \hline
       \rowcolor{Lavender!40}
       3&Copenhagen & High & Cold & DK  &\\
       \hline
       \rowcolor{SkyBlue}
    4&Berlin     & Very High & Mild & DE &\\
    \hline
    \rowcolor{SkyBlue!40}
    5&Aachen     & Low & Mild & DE & \\
    \hline
    \rowcolor{Lavender}
    6&Brussels     & Medium & Mild & BE &\\
    \hline
     \rowcolor{LimeGreen}
    7&Milan     & Medium & Hot & IT &\\
    \hline
    \rowcolor{SkyBlue!40}
    8&Munich & Medium & Mild & DE & \\
       \hline
       \rowcolor{SkyBlue!40}
      9& Paris & Very High & Mild & FR & \\
       \hline
       \rowcolor{LimeGreen!40}
       10&Turin & High & Hot & IT & \\
       \hline
       \rowcolor{SkyBlue}
       11&Seville & High & Hot & ES & \\
       \hline
       \rowcolor{SkyBlue!40}
       12&Valencia & High & Hot & ES & \\
       \hline
    \end{tabular}
    \end{tiny}
    
    \caption{\label{tab:city_properties_sketch_cluster1}Result of the first clustering task, where each color represents a cluster. A light color means that the cluster assignment has been performed by the clustering algorithm.}
    \end{subtable}

\caption{Input sketch, constraints and output sketch of the clustering task.}
\end{table}



\subsection{Sketches for inductive models}
\label{sec:inductive_models}

\begin{table}[h!]
    \centering
\begin{subtable}[t]{1\textwidth}
\centering
    \begin{tiny}
    \begin{tabular}{|c|c|c|c|c|}
    \hline
    June    & July  & Aug   & Total & Profit    \\
    \hline\hline
    610     & 190   & 670   & 1470    & YES       \\
    \hline
    170     & 690   & 520   & 1380  & NO        \\
    \hline
    430     & 350   & ? & ? & ? \\
    \hline
    250     & 650   & ? & ? & ? \\
    \hline
    \end{tabular}
    \end{tiny}
    
    \vspace{1em}
    
    \begin{minipage}{.5\textwidth}

        \centering
        \begin{tiny}
        \begin{tabular}{|c|c|c|c|c|}
        \hline
        June    & July  & Aug   & Total & Profit    \\
        \hline\hline
        \rowcolor{red!33} 610     & 190   & 670   & B4D  & YES       \\
        \hline
        170     & 690   & 520   & 1380  & NO        \\
        \hline
        430     & 350   & ? & ? & ? \\
        \hline
        250     & 650   & ? & ? & ? \\
        \hline
        \end{tabular}
        \end{tiny}

    \end{minipage}%
    \begin{minipage}{.5\textwidth}

        \centering
        \begin{tiny}
        \begin{tabular}{|c|c|c|c|c|}
        \hline
        June    & July  & \cellcolor{blue!33} Aug   & Total & Profit    \\
        \hline\hline
        610     & 190   & \cellcolor{blue!33} 670   & 1470  & YES       \\
        \hline
        170     & 690   & \cellcolor{blue!33} 520   & 1380  & NO        \\
        \hline
        430     & 350   & \cellcolor{blue!33} ? & ? & ? \\
        \hline
        250     & 650   & \cellcolor{blue!33} ? & ? & ? \\
        \hline
        \end{tabular}
        \end{tiny}

    \end{minipage}

    \vspace{1em}

    \begin{minipage}{.5\textwidth}

        \centering
        \begin{tiny}
        \begin{tabular}{|c|c|c|c|c|}
        \hline
        \cellcolor{green!33} June    & July  & \cellcolor{blue!33} Aug   & Total & \cellcolor{green!33} Profit    \\
        \hline\hline
        \cellcolor{green!33} 610     & 190   & \cellcolor{blue!33} 670   & 1470  & \cellcolor{green!33} YES       \\
        \hline
        \cellcolor{green!33} 170     & 690   & \cellcolor{blue!33} 520   & 1380  & \cellcolor{green!33} NO        \\
        \hline
        \cellcolor{green!33} 430     & 350   & \cellcolor{blue!33} ? & ? & \cellcolor{green!33} ? \\
        \hline
        \cellcolor{green!33} 250     & 650   & \cellcolor{blue!33} ? & ? & \cellcolor{green!33} ? \\
        \hline
        \end{tabular}
        \end{tiny}

    \end{minipage}%
    \begin{minipage}{.5\textwidth}

        \centering
        \begin{tiny}
        \begin{tabular}{|c|c|c|c|c|}
        \hline
        \cellcolor{green!33} June    & \cellcolor{green!33} July  & \cellcolor{blue!33} Aug   & Total & Profit    \\
        \hline\hline
        \cellcolor{green!33} 610     & \cellcolor{green!33} 190   & \cellcolor{blue!33} 670   & 1470  & YES       \\
        \hline
        \cellcolor{green!33} 170     & \cellcolor{green!33} 690   & \cellcolor{blue!33} 520   & 1380  & NO        \\
        \hline
        \cellcolor{green!33} 430     & \cellcolor{green!33} 350   & \cellcolor{blue!33} ? & ? & ? \\
        \hline
        \cellcolor{green!33} 250     & \cellcolor{green!33} 650   & \cellcolor{blue!33} ? & ? & ? \\
        \hline
        \end{tabular}
        \end{tiny}

    \end{minipage}

    \vspace{.8em}

    \caption{\label{tab:ice_cream_sales_inductive_models} Illustration of the inductive models sketch.  Top table: simplified ice cream sale numbers.  Middle row: excluding a corrupted row from auto-completion using red (left) and selecting of a column as target using blue (right).  Bottom row: the machine decided to predict August, in blue, from June and Profit, in green (left);  the user improved the system's choice of inputs (right).}
\end{subtable}

\begin{subtable}[t]{.37\textwidth}
    \begin{scriptsize}
    \begin{tabularx}{\textwidth}{X}
    \textbf{Given} the green and blue columns\\
    and the red rows,\\
    \textbf{find} a predictive model  \\
  that predicts the blue column from the green ones, while ignoring the red rows.\\
         \end{tabularx}
         \end{scriptsize}
         \caption{\label{tab:problem_setting_inductive_models} Problem setting of the inductive model learning task. The considered sketch is the bottom left from Table~\ref{tab:ice_cream_sales_inductive_models}}
         \end{subtable}
         \begin{subtable}[t]{.62\textwidth}
         \begin{scriptsize}
         \begin{tabularx}{\textwidth}{X}
         For \textbf{predictor learning}: Launch an autoML instance to learn a model predicting August from June and July, without the first row. The loss function in root mean squared error.\\
         For \textbf{constraint learning}: Learn constraints using June and July to predict August, from the constraint template $\mathbf{S}$\\
         For \textbf{auto-completion}: Use inductive models in the system to predict August from June and July, and learn constraints if none are available and predictors if constraints cannot predict missing values of August.
         \end{tabularx}
         \end{scriptsize}
         \caption{Model learning step solving the problem setting in Table~\ref{tab:problem_setting_inductive_models}}
         \end{subtable}
         
\begin{subtable}[t]{1\textwidth}
\centering
        \begin{tiny}
        \begin{tabular}{|c|c|c|c|c|}
        \hline
        \cellcolor{green!33} June    & \cellcolor{green!33} July  & \cellcolor{blue!33} Aug   & Total & Profit    \\
        \hline\hline
        \cellcolor{green!33} 610     & \cellcolor{green!33} 190   & \cellcolor{blue!33} 670   & 1470  & YES       \\
        \hline
        \cellcolor{green!33} 170     & \cellcolor{green!33} 690   & \cellcolor{blue!33} 520   & 1380  & NO        \\
        \hline
        \cellcolor{green!33} 430     & \cellcolor{green!33} 350   & \cellcolor{blue!33} \textit{460} & ? & ? \\
        \hline
        \cellcolor{green!33} 250     & \cellcolor{green!33} 650   & \cellcolor{blue!33} \textit{540} & ? & ? \\
        \hline
        \end{tabular}
        \end{tiny}
         \caption{Output sketch, where missing values for August have been filled. Predicted values are in italic formatting to indicate that they come from an inductive model. The learned model (constraints, predictor or a combination of both) is stored in the system and is associated with the spreadsheet.}
         \end{subtable}
\caption{Input sketch, problem setting and output sketch of the inductive model learning task.}
\end{table}

In this section, we present the use of sketches for learning and using inductive models for auto-completion. In this context, inductive models refer to predictors, constraints or a combination of the two. 
Learning predictors or constraints typically requires knowing what data to learn from and what is the target to learn.
From this observation, we propose the sketch depicted in Table~\ref{tab:ice_cream_sales_inductive_models}.

First of all, the sketch of Table~\ref{tab:ice_cream_sales_inductive_models} is used to identify target cells and input features.  For instance,  prior to initiating the learning of inductive models, the user might highlight a target column containing empty cells, as in Table~\ref{tab:ice_cream_sales_inductive_models} (middle right). This prompts the system to ignore other empty regions of the spreadsheet, thus focusing the computation to the user's needs and saving computational resources. After a first round of learning, the system might highlight the columns that the value of $August$ was derived from, as in the Table (bottom left). In the example, the system mistakenly used the Profit information to predict the sales for $August$. Although not technically incorrect, as the two values are correlated, this choice does not help in predicting the missing $August$ sales. The user can improve the choice of inputs by de-selecting irrelevant or deleterious inputs and by adding any relevant columns ignored by the system.  A possible result is shown in Table~\ref{tab:ice_cream_sales_inductive_models} (bottom right).

Next, sketches can be used to identify \emph{examples} and \emph{non-examples}.  In Table~\ref{tab:ice_cream_sales_inductive_models} (middle left), the $Total$ is corrupted in one row.  The user can mark that row (e.g., in red) to ensure that the software does neither use it for inferring predictors and constraints nor for making predictions.

In the next paragraphs, we describe how the sketch of Table~\ref{tab:ice_cream_sales_inductive_models} can be used to define a prediction, a constraint learning task and an auto-completion task.


\subsubsection{Prediction}
\label{subsec:prediction}

The task of prediction is one of the most classic tasks in a Data Science process.
Prediction can be decomposed in two steps. First, a predictor is fit on a dataset to predict targets based on input features. Second, the fit model is used to make predictions on new data using similar input features. A common framework to represent these two steps is \emph{fit-predict}, that is for example used in the scikit-learn library~\cite{sklearn_api}. The fit step typically requires input data (also called training data) and target data. The predict step only requires input data.

The prediction sketch depicted in Table~\ref{tab:ice_cream_sales_inductive_models} indicates the input features, the targets and the excluded examples. From the sketch, the prediction task becomes: \textbf{Given} three sets of colored cells, \textbf{find} a predictive model using the columns of the first set of cells to predict the columns of the second set of cells without using rows from the third set of cells.

This prediction task is close to the AutoML task definition~\cite{feurer2015efficient}, with the difference that a loss function usually has to be defined in AutoML. However, we can define default choices for this loss function depending on the type of target feature.
Hence, we can use any AutoML system, such as auto-sklearn~\cite{feurer2015efficient}, TPOT~\cite{OlsonGECCO2016} or auto-WEKA~\cite{kotthoff2017auto} to perform a prediction task given the sketch presented in Table~\ref{tab:ice_cream_sales_inductive_models}.

If the first set of cells is empty,  all columns not in the second set of cells are used as input features.
If the second set of cells is empty, all empty cells are automatically added to the second set. The rationale is that we want to predict all empty cells.

\subsubsection{Learning Constraints and Formulas}

Formulas and constraints are key elements of spreadsheets. 
Formulas are used by users to specify how certain cells can be computed from other cells.
For example, a formula $C_1 = MAX(C_2, ..., C_n)$ specifies that column~$C_1$ is obtained by, for every row, computing the maximum of columns~$C_2$ to~$C_n$.
Constraints can be used to verify whether the data satisfies some invariants and  is consistent.
Simple constraints are often used by spreadsheet users to perform sanity checks on the data~\cite{hermans2013improving}.
For example, a constraint could test whether, in a column~$C_i$, the values are ordered in increasing order.
However, formulas themselves can also be seen as a type of constraints, specifying that the output values correspond to the values computed by the formula.
Therefore, learning constraints and formulas can, in this context, be viewed as simply learning constraints.

In order to assist users in using constraints in their spreadsheets, as well as helping them recover, for example, data exported without formulas from enterprise software packages, existing systems such as \tacle~\cite{kolb2017learning} aim at automatically discovering constraints and formulas in spreadsheets across different tables.
The authors propose a formalization of spreadsheet content into a hierarchical structure of tables, blocks and single rows or columns.
Single rows or columns are denoted as \emph{vectors} to abstract from their orientation and form the minimal level of granularity that constraints can reason about.
This means that a constraint such as $C_1 = MAX(C_2, ..., C_n)$ can only span over entire rows or columns.
Allowing constraints  over subsets of vectors would allow for additional expressiveness at the price of decreased efficiency and a higher risk of finding spurious constraints that are true by accident. 
The data of every table~$T$  is grouped into contiguous blocks of vectors that have the same type and every vector is required to be \emph{type consistent} itself, i.e., all cells within a vector -- and by extension within a block -- need to have the same type.
In practice, these restrictions prohibit blocks or vectors that contain both textual and numeric cells.
Mixed type vectors and blocks will be excluded from the constraint search.
Blocks impose a hierarchy on groupings of vectors through the concept of sub-block containment: a block~$B_1$ is a sub-block~$B_2 (B_1 \sqsubseteq B_2)$ if $B_1$ contains a contiguous subsets of the vectors in~$B_2$.

Similar to Inductive Logic Programming (ILP), constraint learning algorithms  \cite{de2018learning} construct a hypothesis space of possible constraints.
These algorithms then attempt to efficiently search in the hypothesis space for constraints that hold in the example data.
\tacle constructs a hypothesis space using a large catalog of \emph{constraint templates}, e.g., $?_1 = MAX(?_2)$.
This approach is similar to Modelseeker~\cite{beldiceanu2012model}, which uses a catalog of global constraints.
We can now define the tabular constraint problems formally:

    \textbf{Given} a set of instantiated blocks~$\mathbf{B}$ over tables~$\mathbf{T}$ and a set of constraint templates~$\mathbf{S}$, \textbf{find} all constraints $s(B_1', ..., B_n')$ where $s \in \mathbf{S}$, $\forall i: B_i \sqsubseteq B_i' \in \mathbf{B}$ and $(B_1', ..., B_n')$ is a satisfied argument assignment of the template~$s$.

We can use the sketch of Table~\ref{tab:ice_cream_sales_inductive_models} to instruct a constraint learning algorithm to learn constraints for the cells of interest.
Starting from the given tables~$\mathbf{T}$, we can construct a new set of tables~$\hat{\mathbf{T}}$ that contains all colored cells and a minimal number of uncolored cells and no cell colored in red (the third set of colored cells).
This set of tables is computed by collapsing columns and rows that consist solely of uncolored cells and removing cells from the third set of colored cells.
The blocks~$\hat{\mathbf{B}}$ of these tables could be computed by grouping  all neighboring type-consistent vectors.
However, to avoid learning constraints over blocks that are not contiguous in the original tables, vectors that are separated in the original tables~$\mathbf{T}$ by uncolored vectors are not grouped within the same block.
Additionally, to avoid learning constraints over partial rows or columns, only vectors are considered that are subsets {}of vectors that were type-consistent in the original set of blocks~$\mathbf{B}$.
Finally, we can run a tabular constraint learning such as~\tacle on blocks~$\hat{\mathbf{B}}$ to obtain a set of constraints that hold on these cells and can be mapped back to the original tables~$\mathbf{T}$.

We briefly note that, since formulas can also be seen as predictors, and generic constraints -- such as those learned by ModelSeeker~\cite{beldiceanu2012model} or Incal~\cite{ijcai2018-323} -- can also be seen as binary predictors, methods that learn these formulas or constraints can also be used specialized predictors and use the second set of colored cells as to specify output (predicted) columns or rows.

\subsubsection{Auto-completion}

In typical spreadsheet applications, whenever the software detects that the user is entering a predictable sequence of values in a row or column (e.g., a constant ID or a sequence of evenly spaced dates), the remaining entries are filled in automatically.  This is achieved using propagation rules.

This elementary form of \emph{auto-completion}, while useful for automating simple repetitive tasks, is of limited use for data science.  For this reason, we consider a more powerful form of auto-completion, \emph{predictive spreadsheet auto-completion under constraints}, or PSA for short~\cite{kolb2019predictive}.  PSA can be defined as follows: \textbf{Given} a set of tables in a spreadsheet and a set of one or more empty cells, \textbf{find} an assignment of values to the cells.  The key feature of PSA is that the missing values are inferred using one or more predictive models (often classifiers or regressors~\cite{bishop2006pattern}) while ensuring that the predictions are compatible with the formulas and the constraints detected in the spreadsheet.

Let us illustrate predictive auto-completion using the sales data in Table~\ref{tab:ice_cream_sales_inductive_models}.  Some of the values for $August$ are not yet available, hence $Total$ cannot be computed and no conclusion can be drawn about profitability.  Intuitively, PSA auto-completes the table by performing the following steps:
1) find a predictive model for the column $August$ using (some of) the sale numbers for the other months;
2) discover a formula stating that $Total$ is the sum of $June$, $July$, and $August$;
3) find a predictive model for $Profit$ based on both the observed and predicted values;
4) impute all missing cells.

A general strategy for solving PSA was recently proposed that combines two of the core data science tasks considered above, namely learning predictors and learning constraints~\cite{kolb2019predictive}.  This strategy consists of two steps.  In a first step, a set of predictors and formulas for the target cell(s) as well as a set of constraints holding in the data, are learned from the observed portion of the spreadsheet.  Then, the most likely prediction consistent with the extracted constraints is computed.  This is achieved by combining the learned predictors and formulas using probabilistic inference under constraints~\cite{koller2009probabilistic}.  Low-performance models are automatically identified and their predictions are ignored.  The technical aspects of this strategy are outlined in the next Section.

PSA is significantly more useful for interactive data science than standard auto-completion, because it enables non-experts to make use of automatically extracted formulas and constraints without typing them, and to apply predictive models without specifying them.  The assumption is, of course, that an appropriate user interface is available.

From the sketch presented in Table~\ref{tab:ice_cream_sales_inductive_models}, we can derive an auto-completion task, similar to the prediction task described above: \textbf{Given} three sets of colored cells, \textbf{find} a predictive model using the columns of the first set of cells to predict the columns of the second set of cells without using rows from the third set of cells.

In order to solve predictive spreadsheet auto-completion, we rely on \psyche, the implementation introduced in~\cite{kolb2019predictive}.  For the sake of simplicity, we introduce \psyche on the simplest setting, namely auto-completing a single cell.
In PSA, auto-completing a cell amounts to determining the most likely value that is consistent with respect to the constraints holding in the spreadsheet.  If the machine knew what observed cells determine or influence the missing value (e.g., the $August$ sales) and what formulas and constraints hold in the spreadsheet ($Total$ is the sum of $June$, $July$, and $August$), then the problem would boil down to prediction under constraints.  Indeed, one could train a predictive model (e.g., a linear regressor) on the fully observed rows and use it to predict the missing value in the target row.  The caveat is that values that violate the constraints (e.g. the prediction for $August$ might be incompatible with the $Total$ revenue) must be avoided.  In practice, however, no information is given about the relevant inputs and constraints.

To side-step this issue, \psyche extracts a set of candidate predictors and constraints directly from the data.  We discuss this process next.

\paragraph{Solving predictive auto-completion under constraints}  \psyche acquires candidate constraints and formulas from the spreadsheet by invoking TaCLe, a third-party learner specialized for this task~\cite{kolb2017learning}.  As for the predictors, \psyche learns a small ensemble of five to ten models -- including decision trees, linear regressors, or other models -- using randomization.  Since it is unclear which input columns are relevant, each predictor is trained to predict the target value from a random subset of observed columns.  The intuition is that, while most input columns are likely irrelevant, some of the predictors will likely look at some of the relevant ones.  Of course, some of the predictors may perform poorly.  The rest of the pipeline is therefore designed to filter out the bad predictions and retain the good ones.  This is achieved with a combination of probabilistic reasoning and robust estimation techniques, as follows.

First, in order to correct for systematic errors, the outputs of all acquired predictors are calibrated on the training data using a robust estimation procedure.  For example, in class-unbalanced tasks -- like predicting the product ID of a rare ice cream flavour in a sales spreadsheet -- predictors tend to favour the majority class.  The calibration step is designed so to redistribute probability mass from the over-predicted classes to the under-predicted ones.  The calibration is computed using a robust cross-validation procedure~\cite{elisseeff2003leave} directly on the data.  The resulting estimate is further smoothed to prevent over-fitting.  The outcome of this step is a calibrated copy of each base predictor.

In the next step, \psyche combines the calibrated predictions to determine the most likely value for the missing cell.  The issue is that multiple alternatives are available, one for each predictor.  The main goal here is to filter out the bad predictions.  In the simplest case, \psyche performs the combination using a mixture of experts~\cite{jordan1994hierarchical,bishop2006pattern}.  At a high level, this means that each calibrated predictor votes one or more values, where the votes are weighted proportionally on the estimated accuracy of the predictors.  \psyche implements several alternatives which differ in how trust is attributed to the various predictors.  This produces a ranking of candidate values for the target cell.

As a final step, the learned constraints are used to eliminate all invalid candidate values and a winner is chosen.  This guarantees that the value is both valid and suggested by the majority of high-quality (calibrated) predictors.

Auto-completing multiple cells requires  performing the same steps.  The only major complication is that, in this case, since the cells being completed may depend on each other (e.g. $August$, $Total$ and $Profit$ are clearly correlated), \psyche has to find an appropriate order in which to predict them.  Since the rest of the process is intuitively identical to the single-cell case, we do not discuss this further here.  The interested reader can find all the technical details in~\cite{kolb2019predictive}.

\paragraph{Integrating the sketches}  Let us now consider the effect of colored sketches.  So far, we assumed that no information about the inputs, outputs, and constraints is available to the system.  Sketches partially supply this information.  In the previous Section we discussed two types of sketches: 1)~highlighting examples versus non-examples, and 2)~identifying and correcting relevant inputs,  cf. Table~\ref{tab:ice_cream_sales_inductive_models}.  Both can be fit naturally into the design of \psyche and greatly simplify the auto-completion process.

In particular, information about invalid examples enables \psyche to avoid bad predictive models.  The major benefit is that more resources can be allocated to higher-quality models, and that low-quality predictions will be less likely to influence or bias the inference process.  Relevant input information  has similar consequences.

\section{Related Work}

\subsection{Visual Analytics}

Visual analytics refers to  technologies that support discovery by combining automated analysis with interactive visual means \cite{thomas2005illuminating}. \method is therefore tightly linked with visual analytics, as it combines automated data analysis with visual interaction. Visual analytics is typically used to help a user understand or solve a complex problem. Most approaches are tailored to a specific use case or a particular type of data, see \cite{hohman2018visual,amershi2014power,kehrer2012visualization} for  overviews. Some processes of data science have been studied in visual analytics: understanding a machine learning model \cite{krause2016using}, exploring data visualizations \cite{wongsuphasawat2017voyager} or building analysis pipelines \cite{wang2016survey}.
Because these methods are task specific, a challenge in visual analytics is to design interactions that can handle a range of tasks, through different guidance degrees \cite{ceneda2016characterizing}. \method provides one way to use simple interaction through colorings across a range of data science related tasks. \method is therefore a first step towards solving some of the current challenges in visual analytics in the domain of data science.

\subsection{Interactive Machine Learning}
\method also has strong ties with the field of Interactive Machine Learning (IML). IML aims at complementing human intelligence by integrating it with computational power \cite{dudley2018review}. Some of the key challenges of IML are similar to the challenges we are also tackling: inconsistent and uncertain users, intuitive displaying of complex model decisions and wide range of interesting tasks. To solve some of these challenges, most IML approaches focus on a particular type of data: text~\cite{wallace2012deploying}, images~\cite{fails2003interactive}, or time series~\cite{kabra2013jaaba}.  In stark contrast, we focus on spreadsheets, which can store arbitrary combinations of numerical and categorical values, text, and time series.  Moreover, in our setting the task to be solved (e.g., data wrangling, formula extraction or clustering) is not given upfront.  In explorative tasks, the user herself may not know what she is looking for in the data.  Our goal is to help end-users carry out whatever task they have in mind, and which they may have trouble fully articulating.

\subsection{Machine Learning in Spreadsheets}
Small scale user studies about bringing basic machine learning capabilities for non-expert spreadsheet users have been conducted~\cite{Sarkar2014,sarkar2015interactive}. The main conclusion from these studies is that na\"ive end-users are able to successfully use basic machine learning algorithms to predict missing values or assess the quality of existing values. The user can use one button to indicate the data that can be used for learning (the training examples) and another button to apply the learned model to a specific column (the target variable). Visual feedback, in the form of cell coloring or cell annotations is added to communicate with the user. Coloring is used to indicate cells that should be used for training or whether the values imputed by the model are erroneous.
The main difference between these two work~\cite{Sarkar2014,sarkar2015interactive} and \method is that we present a general framework to perform data science tasks using sketches, while these work focus on user studies for the use of colors in spreadsheet for a specific data science task: prediction using k-Nearest Neighbor.


\subsection{Auto-completion and Missing Value Imputation}

Spreadsheet applications often implement simple forms of ``auto-completion'' via propagation rules \cite{gulwani2011automating,harris2011spreadsheet,gulwani2012spreadsheet}.  Clearly, even simple predictive auto-completion is beyond the reach of these approaches.  

Techniques for missing value imputation focus on completing individual data matrices using~\cite{scheuren2005multiple,van2018flexible} using statistics~\cite{van2007multiple} or machine learning~\cite{stekhoven2011missforest}.  These techniques are not designed for spreadsheet data, which usually involves multiple tables, implicit constraints, and formulas.  Several works automate individual elements of the spreadsheet workflow by, e.g., extracting and applying string transformations~\cite{gulwani2011automating,gulwani2015inductive,devlin2017robustfill} and acquiring spreadsheet formulas and constraints hidden in the data~\cite{kolb2017learning}.  Psyche~\cite{kolb2019predictive} combines such tools into a principled predictive auto-completion framework.  In order to do so, it leverages probabilistic inference (using a form of ``chaining''~\cite{van2007multiple})  and learned constraints and formulas to fill in the missing values of multiple related tables.  Psyche is an integral component of the \method interactive data science approach.


\section{Conclusion}

We presented \method, a framework for interactively modeling and solving data science tasks that combines a simple and minimal interaction protocol based on \emph{colored sketches} with \emph{inductive models}.  The sketches enable na\"ive end-users to (partially) define data science tasks such as data wrangling, clustering, and prediction.  At the same time, the inductive models allow the system to clearly capture and reason with general data transformations.  This powerful combination enables even non-experts to solve data science tasks in spreadsheets by \emph{collaborating} with the spreadsheet application.  \method was illustrated through examples on several data science tasks and on concrete use-cases.

Building on \method, an interesting problem is predicting which sketch the user is likely to use given the current state of the spreadsheet.  This is the problem of learning to learn, that is learning what knowledge the user would like to learn.  To do this, an interesting starting point is to observe how users are using sketches to perform the task they have in mind.  Then, learning from these interactions allows us to learn what sketches are typically used in a given state.  Finding suitable representations of such a spreadsheet state is a challenging task, but semantic and structural information, as well as available knowledge are likely to play a key role.

\section*{Acknowledgements} 
This work was funded by  the European Research Council (ERC) under the European Union’s Horizon 2020 research and innovation programme (grant agreement No. 694980) SYNTH: Synthesising Inductive Data Models and by the Research Foundations Flanders and the Special Research Fund (BOF) at KU Leuven through pre- and post-doctoral fellowships for Samuel Kolb.

\bibliographystyle{apalike}
\bibliography{biblio}

\begin{thebibliography}{}

\bibitem[Amershi et~al., 2014]{amershi2014power}
Amershi, S., Cakmak, M., Knox, W.~B., and Kulesza, T. (2014).
\newblock Power to the people: The role of humans in interactive machine
  learning.
\newblock {\em AI Magazine}, 35(4):105--120.

\bibitem[Basu et~al., 2004]{basu2004active}
Basu, S., Banerjee, A., and Mooney, R.~J. (2004).
\newblock Active semi-supervision for pairwise constrained clustering.
\newblock In {\em Proceedings of the 2004 SIAM international conference on data
  mining}, pages 333--344. SIAM.

\bibitem[Beldiceanu and Simonis, 2012]{beldiceanu2012model}
Beldiceanu, N. and Simonis, H. (2012).
\newblock A model seeker: Extracting global constraint models from positive
  examples.
\newblock In {\em International Conference on Principles and Practice of
  Constraint Programming}, pages 141--157. Springer.

\bibitem[Bishop, 2006]{bishop2006pattern}
Bishop, C.~M. (2006).
\newblock {\em Pattern recognition and machine learning}.
\newblock springer.

\bibitem[Bonifati et~al., 2016]{bonifati2016learning}
Bonifati, A., Ciucanu, R., and Staworko, S. (2016).
\newblock Learning join queries from user examples.
\newblock {\em ACM Transactions on Database Systems (TODS)}, 40:24.

\bibitem[Buitinck et~al., 2013]{sklearn_api}
Buitinck, L., Louppe, G., Blondel, M., Pedregosa, F., Mueller, A., Grisel, O.,
  Niculae, V., Prettenhofer, P., Gramfort, A., Grobler, J., Layton, R.,
  VanderPlas, J., Joly, A., Holt, B., and Varoquaux, G. (2013).
\newblock {API} design for machine learning software: experiences from the
  scikit-learn project.
\newblock In {\em ECML PKDD Workshop: Languages for Data Mining and Machine
  Learning}, pages 108--122.

\bibitem[Ceneda et~al., 2016]{ceneda2016characterizing}
Ceneda, D., Gschwandtner, T., May, T., Miksch, S., Schulz, H.-J., Streit, M.,
  and Tominski, C. (2016).
\newblock Characterizing guidance in visual analytics.
\newblock {\em IEEE Transactions on Visualization and Computer Graphics},
  23(1):111--120.

\bibitem[Chambers and Scaffidi, 2010]{chambers2010struggling}
Chambers, C. and Scaffidi, C. (2010).
\newblock Struggling to excel: A field study of challenges faced by spreadsheet
  users.
\newblock In {\em 2010 IEEE Symposium on Visual Languages and Human-Centric
  Computing}, pages 187--194. IEEE.

\bibitem[De~Raedt, 2002]{de2002perspective}
De~Raedt, L. (2002).
\newblock A perspective on inductive databases.
\newblock {\em ACM SIGKDD Explorations Newsletter}, 4(2):69--77.

\bibitem[De~Raedt, 2008]{de2008logical}
De~Raedt, L. (2008).
\newblock {\em Logical and relational learning}.
\newblock Springer Science \& Business Media.

\bibitem[De~Raedt et~al., 2018]{de2018learning}
De~Raedt, L., Passerini, A., and Teso, S. (2018).
\newblock Learning constraints from examples.
\newblock In {\em Thirty-Second AAAI Conference on Artificial Intelligence}.

\bibitem[Devlin et~al., 2017]{devlin2017robustfill}
Devlin, J., Uesato, J., Bhupatiraju, S., Singh, R., Mohamed, A.-r., and Kohli,
  P. (2017).
\newblock Robustfill: Neural program learning under noisy i/o.
\newblock In {\em International Conference on Machine Learning}, pages
  990--998.

\bibitem[Dudley and Kristensson, 2018]{dudley2018review}
Dudley, J.~J. and Kristensson, P.~O. (2018).
\newblock A review of user interface design for interactive machine learning.
\newblock {\em ACM Transactions on Interactive Intelligent Systems (TiiS)},
  8(2):8.

\bibitem[Elisseeff and Pontil, 2003]{elisseeff2003leave}
Elisseeff, A. and Pontil, M. (2003).
\newblock Leave-one-out error and stability of learning algorithms with
  applications.
\newblock {\em NATO science series sub series iii computer and systems
  sciences}, 190:111--130.

\bibitem[Fails and Olsen~Jr, 2003]{fails2003interactive}
Fails, J.~A. and Olsen~Jr, D.~R. (2003).
\newblock Interactive machine learning.
\newblock In {\em Proceedings of the 8th international conference on
  Intelligent user interfaces}, pages 39--45. ACM.

\bibitem[Fayyad et~al., 1996]{fayyad1996kdd}
Fayyad, U., Piatetsky-Shapiro, G., and Smyth, P. (1996).
\newblock The kdd process for extracting useful knowledge from volumes of data.
\newblock {\em Communications of the ACM}, 39(11):27--34.

\bibitem[Feurer et~al., 2015]{feurer2015efficient}
Feurer, M., Klein, A., Eggensperger, K., Springenberg, J., Blum, M., and
  Hutter, F. (2015).
\newblock Efficient and robust automated machine learning.
\newblock In {\em Advances in neural information processing systems}, pages
  2962--2970.

\bibitem[Gulwani, 2011]{gulwani2011automating}
Gulwani, S. (2011).
\newblock Automating string processing in spreadsheets using input-output
  examples.
\newblock {\em ACM Sigplan Notices}, 46(1):317--330.

\bibitem[Gulwani et~al., 2012]{gulwani2012spreadsheet}
Gulwani, S., Harris, W.~R., and Singh, R. (2012).
\newblock Spreadsheet data manipulation using examples.
\newblock {\em Communications of the ACM}, 55(8):97--105.

\bibitem[Gulwani et~al., 2015]{gulwani2015inductive}
Gulwani, S., Hern{\'a}ndez-Orallo, J., Kitzelmann, E., Muggleton, S.~H.,
  Schmid, U., and Zorn, B. (2015).
\newblock Inductive programming meets the real world.
\newblock {\em Communications of the ACM}, 58(11):90--99.

\bibitem[Harris and Gulwani, 2011]{harris2011spreadsheet}
Harris, W.~R. and Gulwani, S. (2011).
\newblock Spreadsheet table transformations from examples.
\newblock {\em ACM SIGPLAN Notices}, 46(6):317--328.

\bibitem[Hermans, 2013]{hermans2013improving}
Hermans, F. (2013).
\newblock Improving spreadsheet test practices.
\newblock In {\em Proceedings of the 2013 Conference of the Center for Advanced
  Studies on Collaborative Research}, pages 56--69.

\bibitem[Hohman et~al., 2018]{hohman2018visual}
Hohman, F.~M., Kahng, M., Pienta, R., and Chau, D.~H. (2018).
\newblock Visual analytics in deep learning: An interrogative survey for the
  next frontiers.
\newblock {\em IEEE transactions on visualization and computer graphics}.

\bibitem[Imielinski and Mannila, 1996]{imielinski1996database}
Imielinski, T. and Mannila, H. (1996).
\newblock A database perspective on knowledge discovery.
\newblock {\em Communications of the ACM}, 39(11):58--64.

\bibitem[Jordan and Jacobs, 1994]{jordan1994hierarchical}
Jordan, M.~I. and Jacobs, R.~A. (1994).
\newblock Hierarchical mixtures of experts and the em algorithm.
\newblock {\em Neural computation}, 6(2):181--214.

\bibitem[Kabra et~al., 2013]{kabra2013jaaba}
Kabra, M., Robie, A.~A., Rivera-Alba, M., Branson, S., and Branson, K. (2013).
\newblock {JAABA: interactive machine learning for automatic annotation of
  animal behavior}.
\newblock {\em Nature methods}, 10(1):64.

\bibitem[Kehrer and Hauser, 2012]{kehrer2012visualization}
Kehrer, J. and Hauser, H. (2012).
\newblock Visualization and visual analysis of multifaceted scientific data: A
  survey.
\newblock {\em IEEE transactions on visualization and computer graphics},
  19(3):495--513.

\bibitem[Kolb et~al., 2017]{kolb2017learning}
Kolb, S., Paramonov, S., Guns, T., and De~Raedt, L. (2017).
\newblock Learning constraints in spreadsheets and tabular data.
\newblock {\em Machine Learning}.

\bibitem[Kolb et~al., 2019]{kolb2019predictive}
Kolb, S., Teso, S., Dries, A., and De~Raedt, L. (2019).
\newblock Predictive spreadsheet autocompletion with constraints.
\newblock {\em Machine Learning}, pages 1--19.

\bibitem[Kolb et~al., 2018]{ijcai2018-323}
Kolb, S., Teso, S., Passerini, A., and Raedt, L.~D. (2018).
\newblock Learning smt(lra) constraints using smt solvers.
\newblock In {\em Proceedings of the Twenty-Seventh International Joint
  Conference on Artificial Intelligence, {IJCAI-18}}, pages 2333--2340.
  International Joint Conferences on Artificial Intelligence Organization.

\bibitem[Koller and Friedman, 2009]{koller2009probabilistic}
Koller, D. and Friedman, N. (2009).
\newblock {\em Probabilistic graphical models: principles and techniques}.
\newblock MIT press.

\bibitem[Kotthoff et~al., 2017]{kotthoff2017auto}
Kotthoff, L., Thornton, C., Hoos, H.~H., Hutter, F., and Leyton-Brown, K.
  (2017).
\newblock Auto-weka 2.0: Automatic model selection and hyperparameter
  optimization in weka.
\newblock {\em The Journal of Machine Learning Research}, 18(1):826--830.

\bibitem[Krause et~al., 2016]{krause2016using}
Krause, J., Perer, A., and Bertini, E. (2016).
\newblock Using visual analytics to interpret predictive machine learning
  models.
\newblock {\em arXiv preprint arXiv:1606.05685}.

\bibitem[Muggleton and De~Raedt, 1994]{muggleton1994inductive}
Muggleton, S. and De~Raedt, L. (1994).
\newblock Inductive logic programming: Theory and methods.
\newblock {\em The Journal of Logic Programming}, 19:629--679.

\bibitem[Muggleton et~al., 1990]{muggleton1990efficient}
Muggleton, S., Feng, C., et~al. (1990).
\newblock {\em Efficient induction of logic programs}.
\newblock Citeseer.

\bibitem[Olson et~al., 2016]{OlsonGECCO2016}
Olson, R.~S., Bartley, N., Urbanowicz, R.~J., and Moore, J.~H. (2016).
\newblock Evaluation of a tree-based pipeline optimization tool for automating
  data science.
\newblock In {\em Proceedings of the Genetic and Evolutionary Computation
  Conference 2016}, GECCO '16, pages 485--492, New York, NY, USA. ACM.

\bibitem[Plotkin, 1970]{plotkin1970note}
Plotkin, G.~D. (1970).
\newblock A note on inductive generalization.
\newblock {\em Machine intelligence}, 5(1):153--163.

\bibitem[Quinlan, 1990]{quinlan1990learning}
Quinlan, J.~R. (1990).
\newblock Learning logical definitions from relations.
\newblock {\em Machine learning}, 5(3):239--266.

\bibitem[Sarkar et~al., 2014]{Sarkar2014}
Sarkar, A., Blackwell, A.~F., Jamnik, M., and Spott, M. (2014).
\newblock {Teach and try: A simple interaction technique for exploratory data
  modelling by end users}.
\newblock In {\em Proceedings of IEEE Symposium on Visual Languages and
  Human-Centric Computing, VL/HCC}, pages 53--56. IEEE Computer Society.

\bibitem[Sarkar et~al., 2015]{sarkar2015interactive}
Sarkar, A., Jamnik, M., Blackwell, A.~F., and Spott, M. (2015).
\newblock Interactive visual machine learning in spreadsheets.
\newblock In {\em 2015 IEEE Symposium on Visual Languages and Human-Centric
  Computing (VL/HCC)}, pages 159--163. IEEE.

\bibitem[Scheuren, 2005]{scheuren2005multiple}
Scheuren, F. (2005).
\newblock Multiple imputation: How it began and continues.
\newblock {\em The American Statistician}, 59(4):315--319.

\bibitem[Stekhoven and B{\"u}hlmann, 2011]{stekhoven2011missforest}
Stekhoven, D.~J. and B{\"u}hlmann, P. (2011).
\newblock Missforest—non-parametric missing value imputation for mixed-type
  data.
\newblock {\em Bioinformatics}, 28(1):112--118.

\bibitem[Thomas, 2005]{thomas2005illuminating}
Thomas, J.~J. (2005).
\newblock {\em Illuminating the path:[the research and development agenda for
  visual analytics]}.
\newblock IEEE Computer Society.

\bibitem[Thornton et~al., 2013]{thornton2013auto}
Thornton, C., Hutter, F., Hoos, H.~H., and Leyton-Brown, K. (2013).
\newblock Auto-weka: Combined selection and hyperparameter optimization of
  classification algorithms.
\newblock In {\em Proceedings of the 19th ACM SIGKDD international conference
  on Knowledge discovery and data mining}, pages 847--855. ACM.

\bibitem[Van~Buuren, 2007]{van2007multiple}
Van~Buuren, S. (2007).
\newblock Multiple imputation of discrete and continuous data by fully
  conditional specification.
\newblock {\em Statistical methods in medical research}, 16(3):219--242.

\bibitem[Van~Buuren, 2018]{van2018flexible}
Van~Buuren, S. (2018).
\newblock {\em Flexible imputation of missing data}.
\newblock Chapman and Hall/CRC.

\bibitem[Van~Craenendonck et~al., 2017]{van2017cobra}
Van~Craenendonck, T., Duman{\v{c}}ic, S., and Blockeel, H. (2017).
\newblock Cobra: a fast and simple method for active clustering with pairwise
  constraints.
\newblock In {\em Proceedings of the 26th International Joint Conference on
  Artificial Intelligence}, pages 2871--2877. AAAI Press.

\bibitem[Van~Craenendonck et~al., 2018]{van2018cobras}
Van~Craenendonck, T., Duman{\v{c}}i{\'c}, S., Van~Wolputte, E., and Blockeel,
  H. (2018).
\newblock Cobras: Interactive clustering with pairwise queries.
\newblock In {\em International Symposium on Intelligent Data Analysis}, pages
  353--366. Springer.

\bibitem[Verbruggen and De~Raedt, 2018]{verbruggen2018automatically}
Verbruggen, G. and De~Raedt, L. (2018).
\newblock Automatically wrangling spreadsheets into machine learning data
  formats.
\newblock In {\em International Symposium on Intelligent Data Analysis}, pages
  367--379. Springer.

\bibitem[Wagstaff et~al., 2001]{wagstaff2001constrained}
Wagstaff, K., Cardie, C., Rogers, S., and Schroedl, S. (2001).
\newblock Constrained k-means clustering with background knowledge.
\newblock In {\em Proceedings of the Eighteenth International Conference on
  Machine Learning}, pages 577--584.

\bibitem[Wallace et~al., 2012]{wallace2012deploying}
Wallace, B.~C., Small, K., Brodley, C.~E., Lau, J., and Trikalinos, T.~A.
  (2012).
\newblock Deploying an interactive machine learning system in an evidence-based
  practice center: abstrackr.
\newblock In {\em proceedings of the 2nd ACM SIGHIT International Health
  Informatics Symposium}, pages 819--824. ACM.

\bibitem[Wang et~al., 2016]{wang2016survey}
Wang, X.-M., Zhang, T.-Y., Ma, Y.-X., Xia, J., and Chen, W. (2016).
\newblock A survey of visual analytic pipelines.
\newblock {\em Journal of Computer Science and Technology}, 31(4):787--804.

\bibitem[Wongsuphasawat et~al., 2017]{wongsuphasawat2017voyager}
Wongsuphasawat, K., Qu, Z., Moritz, D., Chang, R., Ouk, F., Anand, A.,
  Mackinlay, J., Howe, B., and Heer, J. (2017).
\newblock Voyager 2: Augmenting visual analysis with partial view
  specifications.
\newblock In {\em Proceedings of the 2017 CHI Conference on Human Factors in
  Computing Systems}, pages 2648--2659. ACM.

\bibitem[Xing et~al., 2003]{xing2003distance}
Xing, E.~P., Jordan, M.~I., Russell, S.~J., and Ng, A.~Y. (2003).
\newblock Distance metric learning with application to clustering with
  side-information.
\newblock In {\em Advances in neural information processing systems}, pages
  521--528.

\bibitem[Xu and Wunsch, 2005]{xu2005survey}
Xu, R. and Wunsch, D. (2005).
\newblock Survey of clustering algorithms.
\newblock {\em IEEE Transactions on neural networks}, 16(3):645--678.

\end{thebibliography}
\end{document}